\documentclass[conference]{IEEEtran}
\IEEEoverridecommandlockouts
\usepackage{cite}
\usepackage{amsmath,amssymb,amsfonts}
\usepackage{algorithmic}
\usepackage{graphicx}
\usepackage{textcomp}
\usepackage{xcolor}
\usepackage{graphics}
\usepackage{subfigure}
\usepackage{tabularx}
\def\BibTeX{{\rm B\kern-.05em{\sc i\kern-.025em b}\kern-.08em
    T\kern-.1667em\lower.7ex\hbox{E}\kern-.125emX}}

\begin{document}

\title{Back to the Future: Predicting Traffic Shockwave Formation and Propagation Using a Convolutional Encoder-Decoder Network \\
%Traffic Shockwave Propagation Prediction: A Convolutional Encoder-Decoder Approach \\
% {\footnotesize \textsuperscript
% {*}Note: Sub-titles are not captured in Xplore and
}

\author{\IEEEauthorblockN{Mohammadreza Khajeh-Hosseini}
\IEEEauthorblockA{\textit{Department of Civil Engineering} \\
\textit{Texas A\&M University}\\
College Station, USA \\
mreza@tamu.edu}
\and
\IEEEauthorblockN{Alireza Talebpour*}
\IEEEauthorblockA{\textit{Department of Civil Engineering} \\
\textit{Texas A\&M University}\\
College Station, USA \\
atalebpour@tamu.edu}
}

\maketitle

\begin{abstract}
This study proposes a deep learning methodology to predict the propagation of traffic shockwaves. The input to the deep neural network is time-space diagram of the study segment, and the output of the network is the predicted (future) propagation of the shockwave on the study segment in the form of time-space diagram. The main feature of the proposed methodology is the ability to extract the features embedded in the time-space diagram to predict the propagation of traffic shockwaves.
 \end{abstract}

\begin{IEEEkeywords}
Traffic Shockwave Propagation, Convolutional Encoder-Decoder, Traffic Prediction
\end{IEEEkeywords}

\section{Introduction and Background}

The boundary between two different traffic states is known as traffic shockwave. The driving dynamics change from one state to another as the speed of the vehicles and their spacing changes. The differences between the two traffic states can be mild such as a high-speed traffic stream reaching a traffic stream with moderate speed and density, or it can be significant when reaching a high density and low-speed traffic stream (e.g., congested area). In general, when the traffic state changes, the vehicles need to respond by adjusting their speed and acceleration. Currently, the approaches adopted for guidance (e.g., lane changing) of the autonomous vehicles involves the consideration of the current state of the surrounding vehicles in terms of their location and speed with limited attention to the response of the other vehicles to their surrounding environment and how the traffic state could evolve (e.g. formation of shockwaves). As a result, predicting the propagation of traffic shockwave in time and space can help in improving both the safety and performance of the autonomous vehicles. Considering the valuable information that the connected vehicles could provide, we are proposing a methodology to predict the propagation of traffic shockwaves that accounts for both the individual behaviors and the collective change in the state of the traffic.

The state of traffic is characterized by density, flow, and speed that change over time and space (i.e., along the roadway).
% -------------------------------------------------------
% -------------------------------------------------------
% The traffic state prediction approaches differ based on the type of input data, traffic flow model and estimation methodology \cite{seo2017traffic}. 
Lint and Hinsbergen \cite{van2012short} classify traffic prediction methodologies into three general groups of naive, parametric and non-parametric approaches. Naive approaches apply to the methodologies that have no model assumption or parameters driven from data. The naive approaches either assume the future state remains similar to the current state or stays near the average of historical observations for the time of the prediction \cite{eglese2006road}.

Parametric approaches refer to the methodologies that use a traffic flow model with parameters calibrated on historical data or in combination with new observations.
The fundamental diagram in combination with first and second order traffic flow models are among the well studied parametric traffic flow models relating the macroscopic characteristic of traffic state \cite{zheng2016traffic, wang2016multiple}. One of the challenges in the use of parametric models is the trade-off between accuracy and the complexity of traffic prediction models, especially when addressing the time-variant and abnormalities in traffic dynamics. 

Non-parametric approaches, on the other hand, are usually based on simple data-driven methodologies that do not explicitly rely on a traffic flow model. There is a wide range of data analysis and machine learning approaches used in the field of traffic prediction. Some of the common approaches include linear regression \cite{wilby2014lightweight}, different classes of neural network \cite{polson2017deep}, support vector regression \cite{castro2009online}, and time-series forecasting \cite{ kumar2015short}. With the increase in the availability of data resources, the non-parametric approaches have gained more attention in the past decade. Most of the existing non-parametric approaches rely on aggregated and macroscopic traffic data such as flow, density, and speed with limited studies focused on the trajectory of vehicles \cite{elfar2018traffic,khajeh2019two}. With the increase in flow and density, the impact of unexpected driving behaviors on the state of traffic stream increases and more complex dynamics can be observed in the traffic flow. 
% \cite{ossen2008longitudinal}
Consequently, capturing the interactions among vehicles can lead to a better prediction of traffic and change of states. There is a broad body of literature on traffic state prediction as well as comprehensive paper reviews such as \cite{seo2017traffic} that are recommended for further study to interested readers.
% -------------------------------------------------------
% -------------------------------------------------------

The connected vehicles' technology provides the opportunity to disseminate useful data to share with the drivers, or to complement the vehicle's onboard sensors of autonomous vehicles (AV). The connected vehicles share safety-related data regarding their location and speed with other vehicles and traffic control centers to improve their safety and to optimize their travel time and quality. The connectivity provides the opportunity to monitor the traffic stream and how the traffic evolves over time and space more extensively than the location specific traffic monitoring devices. The individual level location data transmitted by connected vehicles help to construct the time-space diagram. The time-space diagram is a comprehensive representation of the traffic stream without any abstraction or  aggregation. The traffic flow dynamics as well as vehicles' interactions and shockwave formation and propagation are embedded in the time-space diagram.  The objective of this study is to introduce a methodology to predict the propagation of traffic shockwaves on a convolutional encoder-decoder neural networks that can directly utilize the time-space diagram in the traffic prediction process. The key innovation of the proposed methodology is introducing the model that can learn the features embedded in the time-space diagram to predict the propagation of the traffic shockwaves.

\section{Methodology}

The interaction between the vehicles and the traffic state is best captured in the time-space diagram. The time-space diagram is the plot of the trajectory of the vehicles traversing the roadway (i.e., space domain) over time. This plot is comprehensive and provides the location and speed (slope of each trajectory) of the vehicles, as well as the interaction between the vehicles and the traffic state. Besides, the traffic shockwaves, which are the boundary between the different traffic states are evident in the time-space diagram. As a result, this study adopts the time-space diagram as a valuable input and predicts the propagation of shockwave in a time-space diagram format.

\subsection{Time-Space Diagram}
In order to construct the time-space diagram as the input for the prediction, this study assumes a connected vehicles environment. The connected vehicles share their state through wireless communication for safety purposes. It is assumed that each vehicle shares a basic safety message (BSM \cite{sae2016j2735D})

\begin{figure*}
\center
\includegraphics[width=0.75\textwidth]{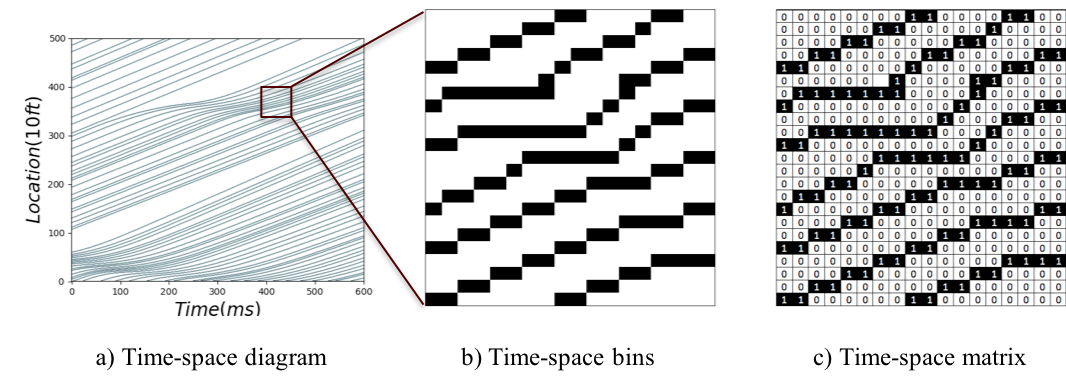}
\caption{Time-Space Diagram \cite{khajeh2019two}}
\label{fig:TS}
\end{figure*}
\noindent transmitted every 100 milliseconds that includes vehicle's speed, location, and heading. In such an environment, the time-space diagram can be generated in the form of a time-space matrix as proposed by Khajeh-Hosseini and Talebpour \cite{khajeh2019two}. The time-space matrix, Fig. \ref{fig:TS}  approximates the time-space diagram by dividing the time and space domains into bins of 10 ft by 100 ms (discretization). The time-space matrix is a binary matrix, and the rows represent the discrete space domain, and the columns represent the discrete time domain. In this matrix, the cell value of one indicates the presence of a vehicle in that space and time bin, and the value of zero indicates an empty bin. The time-space matrix is a 2D tensor that can be used in the convolution process. 

\subsection{Convolutional Encoder-Decoder}
\begin{figure*}
\center
\includegraphics[width=0.85\textwidth]{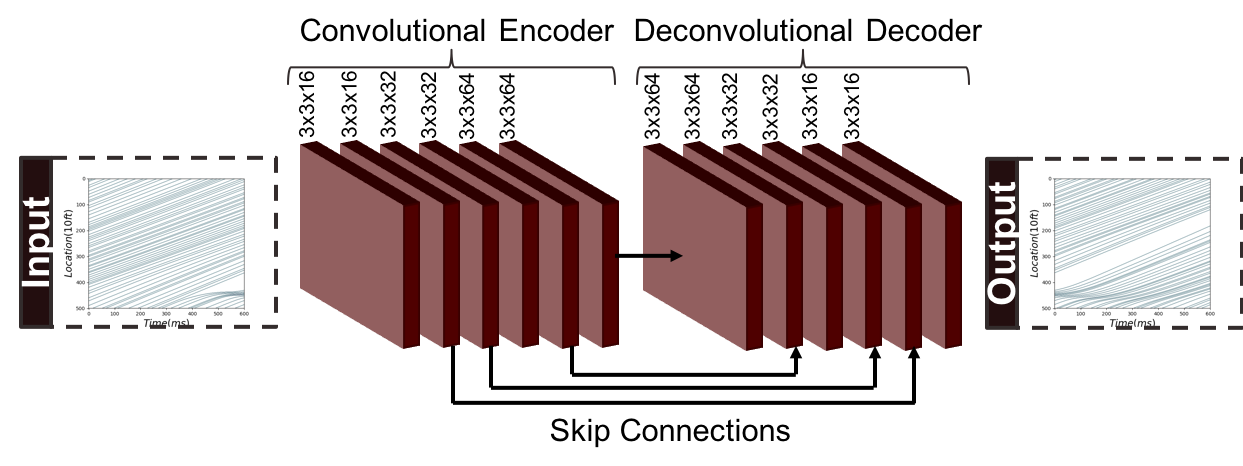}
\caption{Shockwave Prediction: a convolutional encoder-decoder approach.}
\label{fig:encoder-decoder}
\end{figure*}

This study proposes the use of a deep neural network to predict the propagation of the traffic shockwave from the current time-space diagram of the vehicles as shown in Fig. \ref{fig:encoder-decoder}. The convolutional encoder-decoder structure is an appealing type of mapping function for this study as the input and output of the networks are 2D tensors with similar properties. The convolution process uses a receptive window with a fixed size (width, height, and depth) that slides over the neurons of the previous layer. The receptive window provides the opportunity to include the spatial correlation between the nearby units that could construct a feature of interest. Moreover sliding the same filter over the input of the convolution layer ensures a feature identification capability independent of their location. 
% \cite{lecun2015deep}
This location invariant property of the convolution process is one of the main reasons for choosing convolutions to encode the time-space diagram of the vehicles, as the shockwaves in the traffic occur at any point on the time-space diagram. 

There are different convolutional encoder-decoder network architectures depending on the use of convolutional, deconvolutional, pooling and upsampling layers. Some networks only use convolutional layers in both encoder and decoder components such as the  Fully Convolutional Network (FCN) \cite{long2015fully} and Seg-Net \cite{badrinarayanan2017segnet}. While other networks such as DeconvNet \cite{noh2015learning} and RED-Net \cite{mao2016image} use deconvolutional layers in the decoder component. Some of the challenges in the convolutional encoder-decoder network is the vanishing gradient, and reconstructing lost features from the max-pooling and convolution process. The use of skip connections \cite{mao2016image}, and memorizing the maximum features \cite{noh2015learning, badrinarayanan2017segnet} of the pooling process to use for the upsampling %in the decoding process
are the solutions. The skip connections inspired by the residual network (ResNet)\cite{he2016deep}
% and highway network \cite{srivastava2015training}
allow the signal to be propagated to the bottom layers and address the vanishing gradient.

The proposed encoder-decoder architecture in Fig. \ref{fig:encoder-decoder} is inspired by the RED-Net \cite{mao2016image} developed for image restoration, and consists of symmetric layers of convolution and deconvolution with skip-layer connections. The encoder component of the network contains three pairs of convolutional layers with the small receptive window of $3\times3$ and increasing channels from 16 to 64. The decoder component of the network contains symmetrical deconvolutional layers. The deconvolution process, unlike the convolution process, associates a single input to multiple outputs. The encoder component encodes the features embedded in the time-space diagram. The decoder component predicts the propagation of the traffic shockwaves in the form of a new time-space diagram. 

The skip-layers connect the symmetric convolution and deconvolution layers every two layers. The skip-layer connection sums the convolutional feature maps with the deconvolutional feature map element-wise.  The encoding convolutional layers extract the main features and abstracts the input, while the deconvolution layers decode the abstract input and predict the shockwave propagation. The proposed network is deep, and the skip-layer connections provide the opportunity to propagate the gradient to the beginning layers of the network. The skip-layer connections address the vanishing gradient problem in very deep networks. Moreover, the proposed architecture is capable of taking any size of time-space matrix as input since the network only utilizes the convolutional and deconvolutional layers.

%Update the encoder-decoder figure!

\subsection{Data}
\subsubsection{Simulation}
There are limited available data on the trajectory of the individual vehicles to create an extensive data set for the training of the proposed deep neural network that can generalize well. As a result, to construct a comprehensive and large data set, this study uses a microscopic simulator written in the Python programming language to collect the trajectory of the vehicles. The microscopic simulator adopts the Intelligent Driver Model (IDM) \cite{treiber2000congested} as its car-following logic, and the MOBIL \cite{kesting2007general} as its lane-changing logic.

The simulation collects the trajectory of the vehicles traversing a three-lane roadway segment with the length of 40,000 feet over 15 minutes. At every simulation runs, unique and random IDM and MOBIL parameters are assigned to every vehicle to make the simulation more realistic. In order to create a data set with different traffic states from free-flow to fully congested, two types of disturbances are used in the simulation. The first type of disturbance is the speed drop perturbation that forces a random vehicle to decelerate for a small period (e.g., 15 seconds). The second type of disturbance is a slow-moving vehicle for an extended period (e.g., 5 minutes) to create congestion and traffic breakdown. Both of the disturbances result in the formation of shockwaves in the traffic stream. The start time of these two disturbances are limited during periods of $(20i, 20i+20)$ seconds, and $i$ being even numbers between $0$ and $45$. When constructing pairs of input and output data for the training of the model, this limitation becomes useful. The model cannot predict random occurrences of these disturbances, and this limitation helps to exclude the start of these disturbances from the output data. Also, for each simulation run, the desired speed is randomly (uniform distribution) selected from different speed limits including 30, 45, 50, 55, 65, 70, 75 mph to create a more comprehensive data set.

\subsubsection{Input and output data}
\begin{figure}
    \centering
    \includegraphics[width=0.4\textwidth]{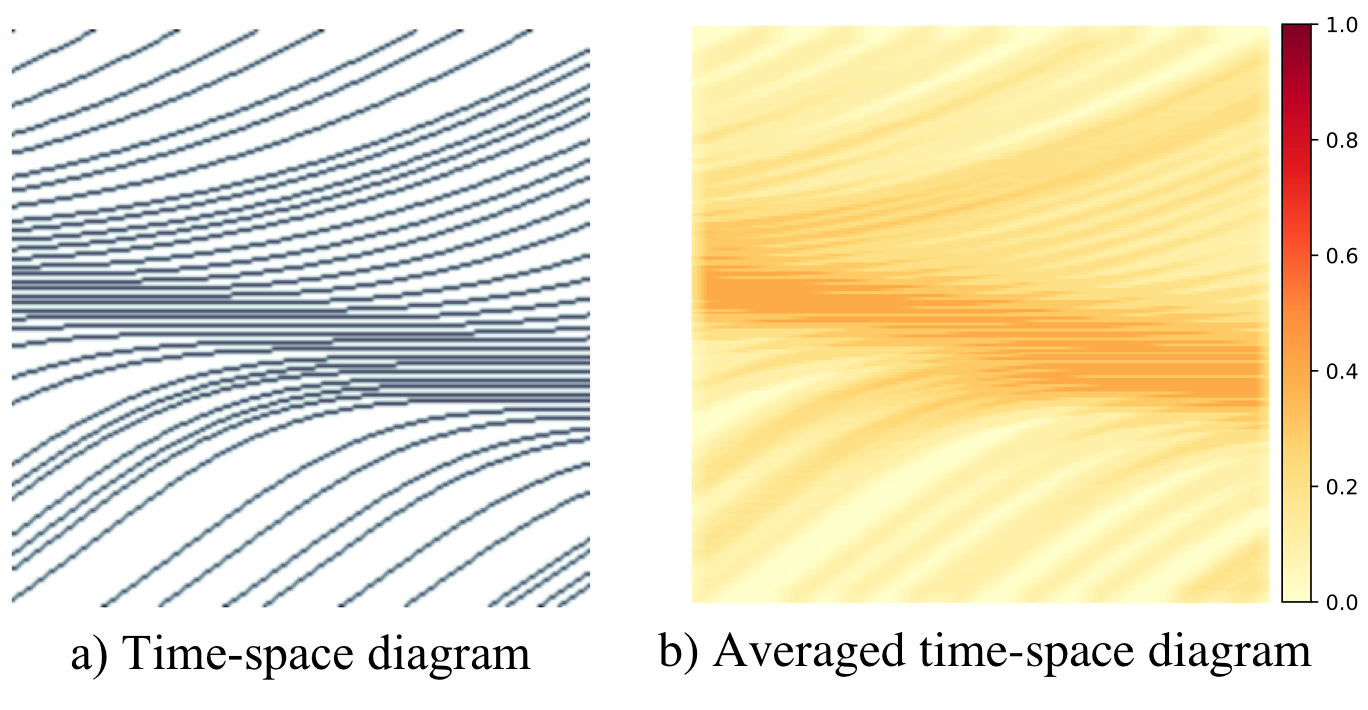}
    \caption{Averaged time-space diagram}
    \label{fig:TSA}
\end{figure}

 The proposed encoder-decoder (Fig. \ref{fig:encoder-decoder}) of this study takes the time-space matrix as input and predicts the propagation of the traffic shockwaves in the same time-space matrix form. The binary time-space matrix (Fig. \ref{fig:TS}) not only presents the traffic shockwaves but also depicts the crisp location of the individual vehicles in time and space domains. Training the network to output a binary tensor of shape $(200, 200)$ is challenging, and breaking the binary constraint improves the training.  Averaging the cells of the time-space matrix with their neighbors, Fig. \ref{fig:TSA}a, blurs the exact location of the vehicles on time and space domains; however, averaging over a small window maintains the propagation of traffic shockwaves (Fig. \ref{fig:TSA}b). The colors of points on the averaged time-space diagram presented in Fig. \ref{fig:TSA}b change from light yellow to red proportional to the value of cell ranging from 0 to 1. Taking the averaged time-space matrix as the type of output improves the training of the network. The encoder-decoder network approximates the mapping function from the averaged time-space matrix of segment $x$ over the period of $(t-20, t)$ to the averaged time-space matrix of the segment $x$ over the period of $(t, t+20)$. The averaged time-space matrix is derived by replacing every cell in the time-space matrix with the average of itself and its neighbors up to 100 feet and 1 second (i.e., averaging window of 100 ft by 1 s).

% \begin{figure}
%     \centering
%     \subfigure[Time-space diagram]
%     {
%         \includegraphics[width=0.2\textwidth]{ts1.png}
%         \label{fig:ts1}
%     }
    
%     \subfigure[Time-space diagram]
%     {
%         \includegraphics[width=0.2\textwidth]{tsa1.png}
%         \label{fig:ts1a}
%     }
%     \caption{Averaged time-space matrix}
%     \label{fig:averaged-TS}
% \end{figure}

\subsubsection{Training data}
The microscopic traffic simulator provides a 40000 ft by 900 s time-space diagram for each lane and every simulation run. This diagram can be divided into 900 smaller time-space diagrams for segments of 2000 ft and a shorter period of 20 s. The 900 smaller time-space diagrams are divided into 450 pairs of input and output data. One could extract more pairs of input and output if relax the limitation on keeping the start of the artificial disturbances in the input data. This study collected data from more than 2000 simulation runs resulting in more than 0.9 million data points. The collected data is divided into three groups of training, validation, and testing sets with ratios of $80\%$, $10\%$, and $10\%$ respectively.

\subsection{Training}
Training is the iterative process of adjusting the trainable parameters of the model to gradually minimize the loss function. The convolutional encoder-decoder of this study contains 180,449 trainable parameters. Adopting the small receptive window of $3\times3$, and fully convolutional and deconvolutional layers kept the number of network parameters small. The model parameters are updated in multiple iterations (steps). At each iteration, the loss function is estimated for a batch of data points, and the parameters are adjusted based on their loss gradient times the learning rate (a small constant). The Adam optimizer \cite{kingma2014adam} is a stochastic gradient-based optimizer that is adapted for the training of the network of this study.

\subsubsection{Loss Function}
The prediction model of this study is a regression model that maps the current averaged time-space diagram to the future averaged time-space diagram. The mean squared error (MSE), equation (\ref{eq:MSE}), is a standard performance measure used as the loss function for the training of regression type neural networks. The output of network (model) $F$ with parameters $\Theta$ for input $X^i$ is $F(X^i;\Theta)$, and the true value of output is $Y^i$. Mean absolute error (MAE), equation (\ref{eq:MAE}), is another performance measure for regression problems. However, the MAE is not useful as the loss function and estimation of gradients in neural networks.

\begin{equation}
    MSE = \frac{1}{n}\sum_{i=1}^{n}||F(X^i;\Theta)-Y^i||^2
    \label{eq:MSE}
\end{equation}

\begin{equation}
    MAE = \frac{1}{n}\sum_{i=1}^{n}||F(X^i;\Theta)-Y^i||
    \label{eq:MAE}
\end{equation}

The input and output of the model of this study are 2D tensors of size $(200, 200)$. 
%, Also, the neighboring cell of the averaged time-space matrix are correlated.
A smoothed version of the output can be constructed by replacing each cell of the output tensor with the average of itself and its neighboring cells. A well trained neural network model is expected to predict outputs very comparable to the true outputs. Besides, it is expected that the smoothed versions of the predicted and true outputs are also comparable. In order to speed up the training (convergence) of the model and to guide the gradients, this study proposes the use of the custom loss function of equation (\ref{eq:loss}). The $MSE_{10}$, $MSE_5$, and $MSE_3$ are the estimated MSE between the true and predicted outputs when smoothed with sliding average windows of size $10\times10$, $3\times3$, and $5\times5$ respectively. The size of the sliding window indicates the extent of neighboring cells considered in the estimation of the average for that cell. Adopting this custom loss function significantly improved the convergence of the training process. 

\begin{equation}
    loss = MSE + 1000(MSE_{10}+MSE_5+MSE_3)
    \label{eq:loss}
\end{equation}

The training process of the model is conducted in two steps. In the first step, the model is trained using the loss function of equation \ref{eq:loss} until the loss value on the validation set started increasing. In the second step, the model is retrained using the MSE, equation (\ref{eq:MSE}), as the loss function. 

\section{Results and Discussions on Results}\label{AA}

In the training process of the model, a batch size of 60 and the early stopping policy are used to prevent overfitting the training data. The loss function is estimated at the end of every epoch (a complete iteration over the entire dataset), and the training is stopped after five epochs from the one with the minimum loss on the validation dataset. Table \ref{tab:training} presents the performance of the network of this study in prediction on the validation and testing dataset. Training the model with the custom loss function, equation (\ref{eq:loss}),  helped the convergence in the first step of training. Also, retraining the model in the second step by adopting the original MSE as the loss function further improved the performance of the model from MSE error of 0.0037 to 0.0029. According to table \ref{tab:training}, the performance of the fully trained model in terms of MSE and MAE on the testing dataset are 0.0030 and 0.0408 respectively.

\begin{table}[]
    \centering
    \caption{Model performance on time-space matrix}
    \begin{tabular}{|c|c|c|}
        \hline
        \multicolumn{3}{|c|}{\textbf{Validation Dataset}}  \\\hline
        & \textbf{$MSE$} & \textbf{{$MSE_{10}+MSE_5+MSE_3$}}\\\hline
        Training Step 1 & 0.0037 & 0.0071 \\\hline
        Training Step 2 & 0.0029 & NA \\\hline
        \multicolumn{3}{|c|}{\textbf{Testing Dataset}}  \\\hline
        % & \multicolumn{2}{|c|}{$MSE$} \\\hline
        & $MSE$ & $MAE$ \\\hline
        % Trained Model & \multicolumn{2}{|c|}{0} \\\hline
        Trained Model & 0.0030 & 0.0408 \\\hline
    \end{tabular}
    \label{tab:training}
\end{table}

\subsection{Traffic shockwave propagation prediction}
\begin{figure}
    \centering
    \includegraphics[width = 0.5\textwidth]{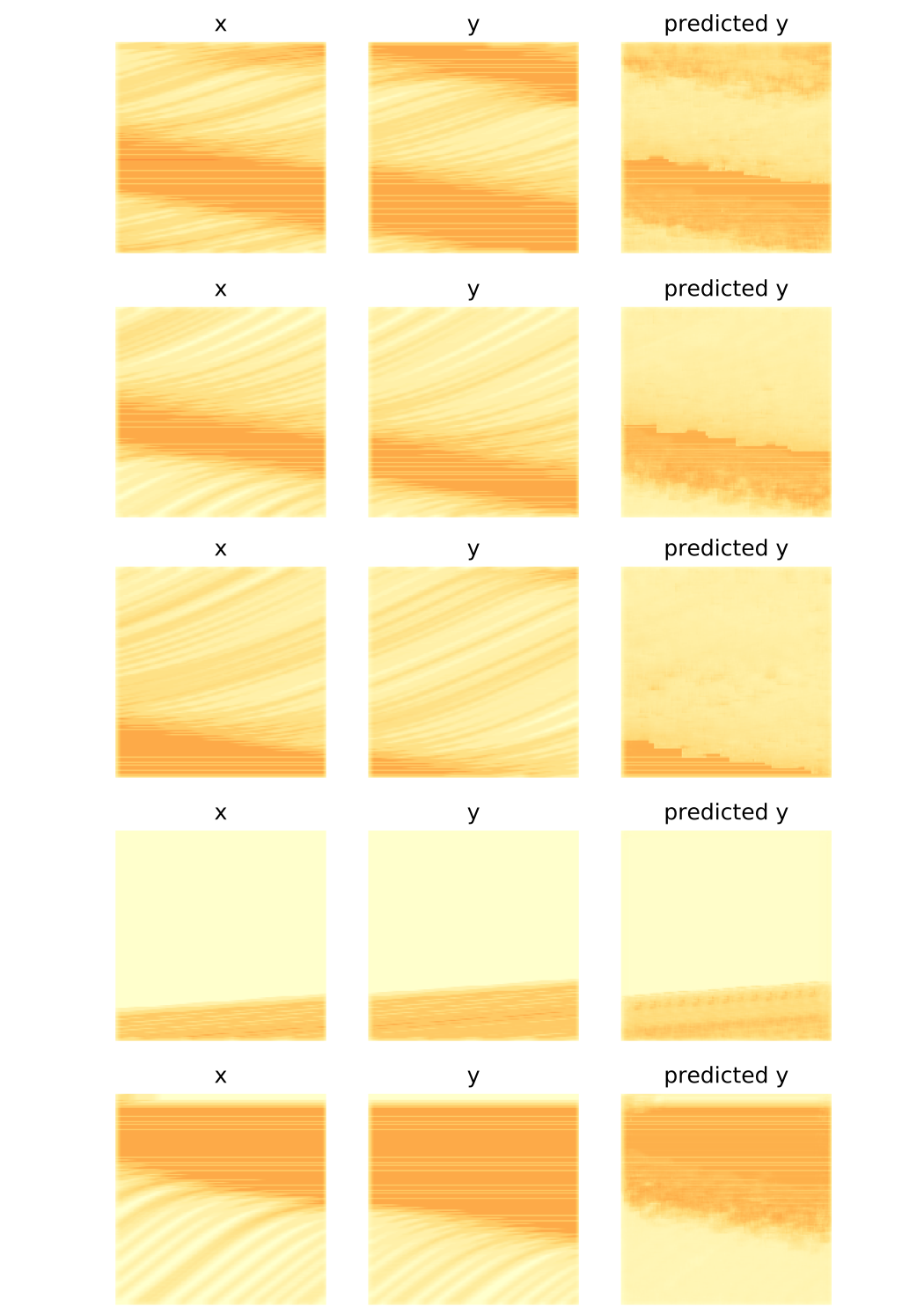}
    \caption{Traffic shockwave propagation prediction results}
    \label{fig:results}
\end{figure}

Fig. \ref{fig:results} presents some of the traffic predictions of the model in the form of the averaged time-space matrix. The prediction model of this study takes the averaged time-space matrix of time $(t-20, t)$ as input (x), and predicts the future averaged time-space matrix of time $(t, t+20)$ as the output (y). Comparing the predictions of the model (predicted y) and the true states of the traffic (y) in Fig. \ref{fig:results}, the model is capable of predicting the propagation of the traffic shockwaves. According to this figure, the predicted averaged time-space diagrams presents dissemination, propagation, the forward and backward movement of the traffic shockwaves over the evaluated segment of roadway.

\subsection{Density time-space matrix}
The traffic shockwave propagation prediction of the network can be evaluated more quantitatively. The traffic shockwave is the boundary between two states of the traffic. Edie \cite{edie1961car} estimates the average density $k(A)$ for a time-space block of $A$ (e.g. 100 feet by 1 second) based on equation (\ref{eq:density}). In this equation, $|A|$ is the area of the time-space block $A$, and $t(A)$ stand for the total time spent by all the vehicles going through block $A$.

\begin{equation}
    k(A) = \frac{t(A)}{|A|}
    \label{eq:density}
\end{equation}

As specified in the methodology section, the time-space matrix is a binary matrix constructed by dividing time and space domains into bins of 10 ft by 100 ms. In this matrix, one represents the presence of a vehicle in that time and space bin, and zero represents an empty bin. The number of occupied bins of the time-space block $A$ is equal to the summation of all the bins of its representative binary time-space matrix (i.e., $sum(A)$). As a result, the total time spent by all the vehicles going through any arbitrary time-space block of $A$ is equal to multiplying the number of occupied bins in that block by 0.1 second, equation \ref{eq:totalTime}.

\begin{equation}
    t(A) = 0.1\times sum(A)
    \label{eq:totalTime}
\end{equation}

Considering Edie's \cite{edie1961car} definition on the average density of a time-space block, the averaged time-space matrix ($\overline{TS}$) can be used to estimate the density time-space matrix ($K$). Similar to the time-space matrix, the rows of this matrix represent the discrete space domain, and their columns represent the discrete time domain. The values of each cell in the matrix $K$ is the average density of a time-space block (e.g., 100 ft by 1 s) centered at that location in time and space. The density time-space matrix depicts the change in traffic state over time and space and consequently the propagation of the traffic shockwaves.

As mentioned in the methodology section, the averaged time-space matrix ($\overline{TS}$) of this study is estimated by replacing every cell in the time-space matrix with the average of itself and its neighbors up to 100 ft and 1 s. Each cell in the time-space matrix is representative of a bin with dimensions of 10 ft in space domain and 0.1 s in the time domain. The averaging window of 100 ft by 1 s is equivalent to a $10\times10$ averaging window on the time-space matrix, in other words, each cell of the averaged time-space matrix ($\overline{TS}$) is the average of 100 cells in the time-space matrix. Therefore, if the averaged time-space matrix ($\overline{TS}$) is multiplied by 100, the cells of the resulting matrix indicate the number of occupied cells in the blocks of 100 ft by 1 s centered on that location on the time-space matrix. As a result, the density time-space matrix ($K$) can be estimated based on equation (\ref{eq:K}). In this equation, the constant 5280 is applied for the unit conversion from feet to mile. Vehicles per mile (vpm) is the unit for the values in the resulting density time-space matrix ($K$).

\begin{equation}
    K = \frac{t(A)}{|A|} = \frac{100 \times\overline{TS}\times0.1}{100 \times 1}\times5280 = 528 \times \overline{TS}
    \label{eq:K}
\end{equation}

According to equation (\ref{eq:K}), the averaged time-space matrix ($\overline{TS}$) can be converted to the density time-space matrix ($K$) by a constant scalar. Therefore, the prediction (output) of the model is proportional to the density time-space matrix. The performance of the model in table \ref{tab:training} are updated for the density time-space matrix presented in table \ref{tab:densityTraining}. According to table \ref{tab:densityTraining}, the mean absolute error of the model in predicting the density for small blocks of 100 ft by 1 s is 21.54 vpm. Root mean squared error (RMSE) is another valuable performance measure that has the same unit as the output. Based on table \ref{tab:densityTraining}, the RMSE of the model in the prediction of the density on the testing dataset is estimated as 28.91 vpm. Considering a range of 200 vpm for the density, the MAE and RMSE of the model are between \%10 to \%14 of the range of density.

\begin{table}[]
    \centering
    \caption{Network performance on density time-space matrix}
    \begin{tabular}{|c|c|c|}
        \hline
        \multicolumn{3}{|c|}{\textbf{Validation Dataset}}  \\\hline
        & \textbf{$MSE$} & \textbf{{$MSE_{10}+MSE_5+MSE_3$}}\\\hline
        Training Step 1 & 1031.50 & 1979.36 \\\hline
        Training Step 2 & 808.47 & NA \\\hline
        \multicolumn{3}{|c|}{\textbf{Testing Dataset}}  \\\hline
        % & \multicolumn{2}{|c|}{$MSE$} \\\hline
        & $MSE$ & $MAE$ \\\hline
        % Trained Model & \multicolumn{2}{|c|}{0} \\\hline
        Trained Model & 836.35 & 21.54 \\\hline
    \end{tabular}
    \label{tab:densityTraining}
\end{table}

%Prove that these averaged time-space matrix can be also converted to averaged flow and density based on Eddie's definition

\section{Conclusion}
This study proposes a methodology to predict the propagation of traffic shockwaves in the form of the averaged time-space matrix. The averaged time-space matrix is comparable to a density time-space matrix derived from Edie's definition of average density. The traffic shockwave is the boundary between two traffic states, and the density time-space matrix depicts the state changes in the form of density. The result of the analysis indicated that the model is capable of predicting the dissemination, propagation, the forward and backward movement of the traffic shockwaves over the study segment. Moreover, the performance of the model in the form of MAE and RMSE for predicting the density time-space matrix is 21.54 vpm and 28.91 vpm respectively. Considering a range of 0 to 200 vpm for the density, the performance of the model is acceptable for prediction of the traffic shockwaves propagation.

\section{Acknowledgment}
This material is based upon work supported by the National Science Foundation under Grant No. 1826410.


% \end{thebibliography}
\bibliography{references.bib}
\bibliographystyle{IEEEtran}

\end{document}